\documentclass{article}

    \PassOptionsToPackage{numbers, compress}{natbib}
\usepackage[final]{neurips_2024}

\usepackage[utf8]{inputenc} % allow utf-8 input
\usepackage[T1]{fontenc}    % use 8-bit T1 fonts
\usepackage{hyperref}       % hyperlinks
\usepackage{url}            % simple URL typesetting
\usepackage{booktabs}       % professional-quality tables
\usepackage{amsfonts}       % blackboard math symbols
\usepackage{nicefrac}       % compact symbols for 1/2, etc.
\usepackage{microtype}      % microtypography
\usepackage{xcolor}         % colors

\usepackage{wrapfig}

\usepackage{amsmath,amsfonts}
\usepackage{algorithmic}
\usepackage{graphicx}
\usepackage{textcomp}
\usepackage{array,colortbl,xcolor}

\usepackage{tkz-graph}

\usepackage{pgfplots}
\usepackage{tikz}
\usetikzlibrary{decorations.pathreplacing,calligraphy}
\usetikzlibrary{patterns}
\usetikzlibrary{arrows}
\usetikzlibrary{calc}
\usetikzlibrary{shapes}

\usepackage{floatrow}

\usepackage{pgfkeys}
\usetikzlibrary{backgrounds}
\pgfplotsset{compat=newest}
\usepackage[export]{adjustbox}
    
\usepackage{comment}
    
\usepackage{xspace}

\usepackage{algorithm}
\usepackage{algorithmic,multicol}
\usepackage{multirow}

\usepackage{booktabs}

\usepackage{tabularx}

\usepackage{pdfpages}

\usepackage{listings}

\usepackage{dsfont}

\usepackage{caption}
\usepackage{subcaption}

\title{\hddebug: Debugging TinyML models on-device using Hyper-Dimensional computing}

\author{%
  Nikhil P Ghanathe\\
  The University of British Columbia\\
  \texttt{nikhilghanathe@ece.ubc.ca} \\
  \And
  Steven J E Wilton \\
  The University of British Columbia \\
  \texttt{stevew@ece.ubc.ca} \\
}

\newcommand{\hddebug}{\textsc{Debug-HD}\xspace}

\definecolor{bblue}{HTML}{4F81BD}
\definecolor{rred}{HTML}{C0504D}
\definecolor{ggreen}{HTML}{9BBB59}
\definecolor{ppurple}{HTML}{9F4C7C}
\definecolor{yyellow}{HTML}{DDAE06}

\tikzstyle{bar1} = [bblue, fill=bblue, postaction={pattern=north east lines}]
\tikzstyle{bar2} = [rred, fill=rred]
\tikzstyle{bar3} = [ggreen, fill=ggreen, postaction={pattern=north west lines}]
\tikzstyle{bar4} = [ppurple, fill=ppurple, postaction={pattern=crosshatch dots}]
\tikzstyle{bar5} = [yyellow, fill=yyellow, postaction={pattern=grid}]

\newcommand{\graphHeight}{0.15\textheight}

\begin{document}

\maketitle

\begin{abstract}
% TinyML models often operate in remote, dynamic environments without any cloud connectivity, making them prone to failures. In this scenario, ensuring reliability requires not only detecting model failures but also identifying their root causes. However, transient failures and privacy concerns and inability to interrupt application complicate the use of raw sensor data for offline debugging. This highlights the need for an intelligent on-device debugging system, which is particularly challenging for KB-sized TinyML devices due to their limited memory, computational power, and energy constraints.
% We propose \hddebug, a novel resource-efficient approach for on-device debugging that utilizes hyper-dimensional computing (HDC) to address these resource limitations. Our method introduces a new HDC encoding technique that leverages conventional neural networks, enabling \hddebug to outperform prior binary HDC methods by 27\% on average in detecting input image corruptions across various image and audio datasets. 

TinyML models often operate in remote, dynamic environments without cloud connectivity, making them prone to failures. Ensuring reliability in such scenarios requires not only detecting model failures but also identifying their root causes. However, transient failures, privacy concerns, and the safety-critical nature of many applications—where systems cannot be interrupted for debugging—complicate the use of raw sensor data for offline analysis. 
% This highlights the need for an intelligent on-device debugging system, which is particularly challenging for KB-sized TinyML devices due to their limited memory, computational power, and energy constraints.
We propose \hddebug, a novel, resource-efficient on-device debugging approach optimized for KB-sized tinyML devices that utilizes hyper-dimensional computing (HDC). Our method introduces a new HDC encoding technique that leverages conventional neural networks, allowing \hddebug to outperform prior binary HDC methods by 27\% on average in detecting input corruptions across various image and audio datasets.
% This makes it an essential tool for enhancing the reliability of TinyML devices.

% We propose \hddebug, a novel resource-efficient approach for on-device debugging using hyper-dimensional computing (HDC), which helps combat the resource scarcity on tinyML devices. Our method introduces a new HDC encoding technique leveraging conventional neural networks, which leads to \hddebug also outperforms previous binary HDC methods in detecting input image corruptions on various image classification datasets.

% achieving a 31\% average accuracy improvement over Vanilla HDC. 
% \hddebug also outperforms previous binary HDC methods in detecting input image corruptions on various image classification datasets.
\end{abstract}

\section{Introduction}
\label{sec:intro}

Recent advances in machine learning (ML) and embedded systems have enabled ML to run on KB-sized, milliwatt-powered devices known as \textit{TinyML}. These \textit{always-on} devices perform all computations locally without any cloud connectivity~\cite{tinyml}. TinyML has increasingly been used in mission-critical scenarios such as autonomous navigation~\cite{vargas2021-overview-av-vulnerability} and healthcare diagnostics~\cite{gesturepod}.  
% Many tinyML applications are safety-critical (e.g., autonomous navigation, healthcare diagnostics) and deployed in 
However, when deployed in uncertain environments, where inputs can become unpredictably \textit{corrupted} (e.g., sensor failure or weather changes)~\cite{kang2018-model-assertions}, maintaining model \textit{reliability} becomes crucial, especially when device access is limited.
% \begin{wrapfigure}{r}{0.4\textwidth}
% \begin{figure}
%     \centering
%     \begin{subfigure}[]{0.55\textwidth}    
%     \hspace{-1cm}
%     \scalebox{0.7}{
%     \includegraphics[width=1\textwidth]{figs/system-overview.pdf}
%     }
%     \hspace{-1cm}
%     \captionsetup{justification= centering, singlelinecheck = false}
%     \caption{}
%       \label{fig:hydeit}
%     \end{subfigure}
%     \begin{subfigure}[]{0.3\textwidth}    
%         \centering
%         \scalebox{0.7}{
%         \input{figs/acc-v-hyper-d}
%         }
%         \captionsetup{justification= centering, singlelinecheck = false}
%         \caption{}
%         \label{fig:acc-v-hyper-d}
%     \end{subfigure}
%     \caption{Figure~\ref{fig:hydeit} provides the overview of training (solid lines) and inference (dotted lines) of \hddebug. Figure~\ref{fig:acc-v-hyper-d} plots the top-1 accuracy of a Vanilla HDC classifier for hyper-dimension values from 200-10k on the task of identifying corruptions in CIFAR10}
% \end{figure}
% \end{wrapfigure}
% A model is considered \textit{reliable} if its failures can be detected, and the root cause of those failures can be identified and addressed effectively. 
% A \textit{reliable} tinyML system deployed in the field must have two components:~1)~\textit{model monitor} to detect drop in model performance 2)~\textit{model debugger} to identify the source of the failures. However, both incur a significant overhead on an already resource-scare tinyML device. 
A model is considered \textit{reliable} if its failures can be detected, and the root cause of those failures can be identified and addressed effectively. While recent works~\cite{ghanathe2024-qute} have proposed resource-efficient monitoring mechanisms for detecting failures in tinyML systems, ensuring reliability often requires identifying these root causes, sometimes \textit{in the field}.  
However, this is challenging on KB-sized tinyML devices, often running bare-metal applications, as transmitting raw sensor data for offline diagnosis can raise privacy and regulatory concerns~\cite{chauduri2024-tinyml-healthcare-challenges}, especially without secure communication hardware. Moreover, many tinyML applications are safety-critical (e.g., remote patient monitoring~\cite{ahmed2022-tinycare}, autonomous navigation~\cite{loquercio2018-dronet}), and thus cannot be interrupted for transmitting data~\cite{dias2018-wearable-healthcare-challenges}. In such scenarios, on-device debugging becomes crucial to ensure reliability, offering advantages such as 1)~triggering \textit{corrective actions} by identifying the root cause of failures~\cite{kang2020-model-assertions}, and 2)~enhancing \textit{active learning} by uncovering \textit{hard-to-obtain} yet highly informative samples~\cite{settles2009-active-learning-lit-survey, coleman2020-active-learning-data-selection}.

Unlike most prior works on on-device debugging~\cite{holanda2019-overlay-mldebug, qiu2022-mlexray,hao2023-mldebug-mobile}, we focus on \textit{What input changes caused model failure?} instead of \textit{Why a model fails for given inputs?} Thus, we focus on \textit{diagnosing failures} in field-deployed models in this work, the first crucial step in debugging.
To combat resource scarcity, we investigate hyper-dimensional computing (HDC)~\cite{kanerva2009-hdc} as a resource-efficient paradigm for on-device debugging in TinyML, leveraging HDC's lightweight operations and memory-optimized representations. To explore this idea, we focus on the task of identifying the source of \textit{corruptions} in input images (e.g., noise, weather, blur etc.) that cause model accuracy to drop. We propose \hddebug, a novel binary HDC classifier that can effectively classify these corruptions (data shifts) in real-time. In addition, we address the limitations of HDC in the tinyML context by formulating an efficient encoding method with the assistance of a conventional neural network (NN), thereby allowing \hddebug to outperform all prior binary HDC works even in low hyper-dimensions (hyper-d). \hddebug achieves 12\% higher accuracy on average than the best-performing prior work on various datasets. To the best of our knowledge, we are among the first to explore on-device debugging in remote, field-deployed tinyML devices. We believe that the proposed work offers developers a practical solution for diagnosing and addressing model failures, enhancing autonomy, reliability, and real-time diagnostic capabilities in tinyML.

\begin{wrapfigure}{r}{0.4\textwidth}
    \vspace{-1.3cm}\scalebox{0.9}{
    \includegraphics[width=1.\textwidth]{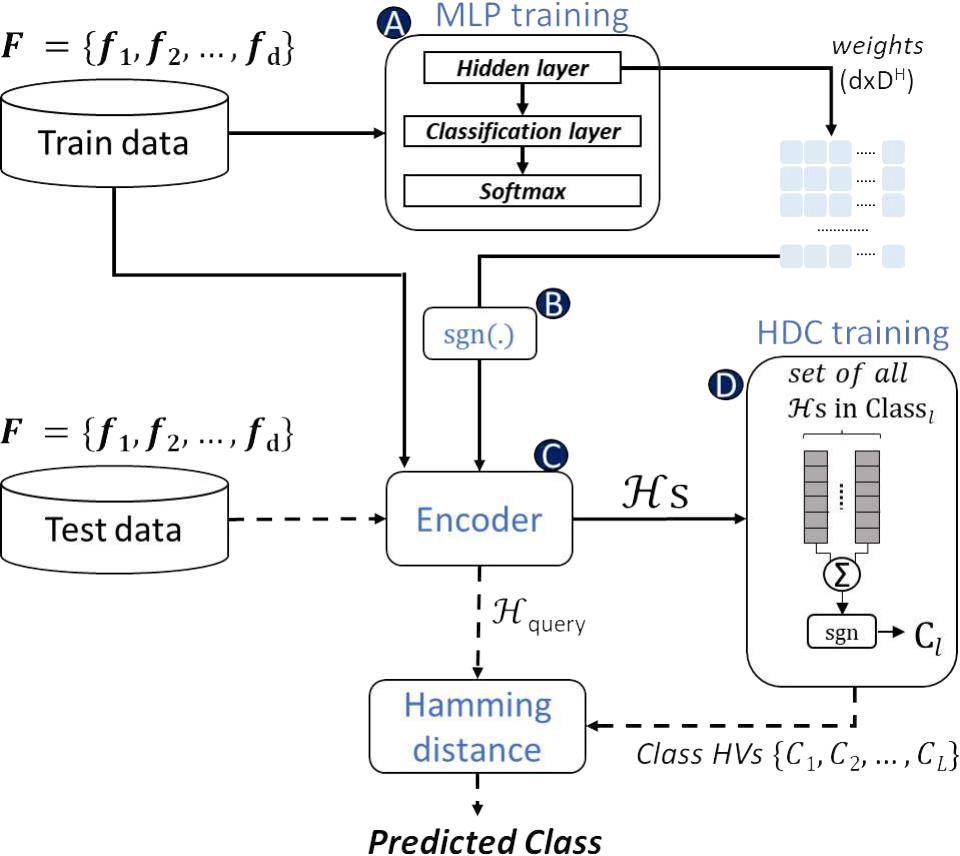}
    }
    \hspace{-1cm}
    % \captionsetup{justification= centering, singlelinecheck = false}
    \vspace{-2mm}
    \caption{\hddebug overview with training (solid lines) and inference (dotted lines) flows\vspace{-2mm}}
    \label{fig:hddebug}
\end{wrapfigure}

\section{Background and Preliminaries}
\label{sec:background}
\textbf{Traditional ML debug}~
TensorFlow Debugger (\textit{tfdbg})~\cite{tfdbg} provides deep insights into TensorFlow graphs for software-like debugging, while TensorBoard~\cite{tensorboard} and TensorWatch~\cite{tensorwatch} allow visual monitoring of model statistics. Typically, lossy compression (e.g., statistical summaries) is sufficient for machine learning debugging. Tools like ML-Exray~\cite{qiu2022-mlexray} catch deployment issues (e.g., quantization/preprocessing bugs), while Nazar~\cite{hao2023-mldebug-mobile} performs root-cause analysis. U-TOE~\cite{huang2023-utoe} and RIOT~\cite{huang2024-riot} add over-the-air (OTA) debugging capabilities. Other debugging techniques~\cite{kang2018-model-assertions, cidon2021-characterizing-model-instability} focus on model assertions and characterizing model instability. The closest works on on-device ML debug~\cite{holanda2019-overlay-mldebug, holanda2021-flexible-fpga-mldebug, onchip-debug-holanda2019} track statistics like sparsity and magnitude distributions.
Another line of research focuses on \emph{interpretability/explainability} methods~\cite{ali2023-xai-survey, adadi2018-xai-survey-1, kim2018-CAV, abid2022-counterfactual_CAV, sundararajan2017-axiomatic-dnn, baehrens2010-explain-classification, zeiler2014-deconvnet, pieter2018-patternet, springenberg2014-guided-backprop, jamil2023-advanced-gradCAM}, which help explain the inner workings of a ML model in a human-understandable way. However, almost all prior works assume easy hardware access, which is often impractical for field-deployed tinyML systems.

\subsection{HDC preliminaries}
\label{subsec:hdc_prelim}
HDC \textit{encodes/projects} inputs into hyper-dimensions to exploit the higher \textit{discriminative} properties of hyper-dimensional representations, which in turn enables simpler learning processes that are both compute and memory friendly. There are three key stages in HDC model development (see Figure~\ref{fig:hddebug}). 

\noindent
1)~\textit{Encoding}:~All feature vectors in train set with dimension $d$ are encoded/projected into hyper-dimension $D^H$ using a projection matrix (usually randomly-generated~\cite{chang2023-hdc-survey}) of size $d\times D^H$ such that $D^H>>d$. This is a simple matrix-vector multiplication. The encoded vector is called a \textit{hypervector}.
% rsuch that . Each feature vector with dimension $d$ is projected into the hyperspace with dimension $D_H$ using a randomly-generated projection matrix of size $d\times D_H$. The encoding reduces to a simple matrix-vector multiplication operation. There are several other encoding methods~\cite{chang2023-hdc-survey} in the literature, but the random projection method is widely-used.

\vspace{-1mm}
\noindent
2)~\textit{Training}:~The class hypervectors (HV) are learned by summing all encoded HVs associated with the class. For example, consider a balanced dataset $\mathcal{D}$ with size $|\mathcal{D}|$ and $L$ classes. Let $\{f_{l_1}, f_{l_2}, ... f_{l_N}\}$ be the $N$ samples from $\mathcal{D}$ belonging to class $l$ such that $|\mathcal{D}| = N\times L$. The encoded versions are given by $\{\mathcal{H}_{l_1}, \mathcal{H}_{l_2}, ... \mathcal{H}_{l_N}\}$. 
Then, $\mathcal{C}_l = \sum_{i=1}^{N} \mathcal{H}_{l_i}$. $\{\mathcal{C}_1, \mathcal{C}_2, ..,\mathcal{C}_L$\} is the set of all class HVs.
% Thus, HDC enables single-pass training and even performs well with fewer data points, which is highly beneficial for the edge computing paradigm.

\vspace{-1mm}
\noindent
3)~\textit{Inference}:~The unseen test input is encoded using the same projection matrix as before to create the query HV ($\mathcal{H}_{query}$). Next, the similarity score of $\mathcal{H}_{query}$ is calculated with respect to each class HV and the class with the highest score is predicted. $l_{predicted} =  \underset{{l\in \{1,2,..,L}\}}{argmax}(\delta(\mathcal{H}_{query}, \mathcal{C}_l))$ 
% \begin{eqnarray}
%     l_{predicted} =  \underset{{l\in \{1,2,..,L}\}}{argmax}(\delta(\mathcal{H}_{query}, \mathcal{C}_l))
% \end{eqnarray}

\begin{wrapfigure}{r}{0.3\textwidth}
        \centering
        \vspace{-0.6cm}
        \scalebox{1}{
        \begin{tikzpicture}
	\pgfplotstableread{
        hyper-d  acc
	200		0.3807
        250		0.3862
        300		0.4098
        350		0.4146
        400		0.4196
        500		0.4353
        600		0.4451
        700		0.4598
        800		0.452
        900		0.4785
        1000		0.4722
        2000		0.5016
        3000		0.5045
        4000		0.5079
        5000		0.5162
        6000		0.5186
        7000		0.5203
        8000		0.5273
        9000		0.5191
        10000		0.5226
        }\data
 %        \pgfplotstableread{
 %        hyper-d  acc-vanilla   acc-lehdc
	% 200		0.3807        0.2156
 %        300		0.4098         0.302
 %        500		0.4353         0.3334
 %        700		0.4598         0.4049
 %        1000		0.4722     0.4232   
 %        5000		0.5162     
 %        10000		0.5226
 %        }\datanew
        
		% \begin{axis}
            \begin{semilogxaxis}
                [
			% Chart dimensions
			width=\textwidth,
			height=\graphHeight*2 - 4cm,
			%
			% Ticks config
			% Removing smaller tick lines since grids are used
			major tick length=3pt,
			%
			% Grid lines config
			% Y axis grids are major grids and X axis grids are minior grids
			major grid style={dashed,color=gray!50},
			minor grid style={color=gray!50},
			% ymajorgrids=true,
			% xmajorgrids=true,
% 			xminorgrids=true,
			%
			% X axis config
% 			xtick=data,
% 			xticklabels from table={\data}{ee-placement},
			xticklabel style={text height=5pt,font=\scriptsize},
			%
			% X label config
			xlabel = \text{Hyper-dimension},
			xlabel style={font=\scriptsize},
		    xlabel near ticks,
                xmin = 100,
                xmax = 10000,
                scaled x ticks=false,
                xtick = { 1000, 5000, 10000},
                % extra x ticks={0, 500, 1000, 5000,10000},
                xticklabels={  1k, 5k, 10k},
		      % extra x ticks={0,200, 500, 700, 1000, 5000,10000},
			%extra x tick style={
			%	grid=minor,
		    %	xticklabel=\empty
			%},
			%
			% Y axis config
			ymin=0.36,ymax=0.55,
			yticklabel style={font=\scriptsize},
			%
			% Y label config
			ylabel style={align=center,font=\scriptsize},
			ylabel=Accuracy,
			% ylabel near ticks,
			% Legend config
			% legend style={at={(0.55,1.0)},anchor=south,cells={align=left},draw=none,font=\scriptsize,text width=2.5cm},
               legend style={at={(2,2)},anchor=north east,draw=none, cells={align=left}, font=\scriptsize,text width=0.8cm},
			legend columns=1,
                legend pos= north east,
			%
			% Bar config
			%ybar=3pt,
			%area legend,
			%bar width=3pt,
			enlarge x limits=0.1,
			%
			% Text on each bar config
			% point meta set to explicit since the log scale changes the value of \pgfplotspointsmeta
			%point meta=explicit symbolic,
			%nodes near coords={\textbf{{\pgfplotspointmeta}}},
			%every node near coord/.append style={font=\tiny, rotate=90, anchor=west},
			%nodes near coords align={vertical}
		]
             % Highlight the region from x=0 to x=1000
            % \addplot[fill=yellow, opacity=0.2] coordinates {(0,0.36) (1000,0.36) (1000,0.55) (0,0.55)};
                % Annotation for the highlighted region
            % Highlight the region from x=0 to x=1000
            \addplot[fill=yellow, opacity=0.2] coordinates {(100,0.36) (1000,0.36) (1000,0.55) (100,0.55)};
            % Annotation for the highlighted region
            \node at (axis cs: 300,0.54) [align=center, anchor=north] {\tiny \textbf{Target} \\ \tiny \textbf{$D^H$}};

		\addplot+[bblue, smooth, mark=] table[x=hyper-d, y=acc] from \data;

% 		\addplot +[mark=none] coordinates {(2, -1) (2, 3)};
% 		\addplot [ppurple, mark=*] table[x=ee-placement,y=ee-accuracy] {\data};
% 		\addplot [rred] table[x=X,y=mkr-speed] {\maxscaleData};
		
		%\addplot [bar1] table[x=X,y=uno-speed] {\maxscaleData};
		%\addplot [bar2] table[x=X,y=mkr-speed] {\maxscaleData};
		
		% \addlegendentry{1stack}
  %           \addlegendentry{2stack}
  %             \addlegendentry{3stack}
  %             \addlegendentry{4stack}

% 		\addlegendentry{Reference}

        % \node (resnet) at (anchor.south) [ionode,black,  xshift=0.5cm, yshift=-1.5cm] {a) Resnet};
        
		\end{semilogxaxis}
            % \end{axis}
  
% 		\node (anchor) at (0,0) [ionode,black, xshift=2.7cm, yshift=-5.5] {a) Resnet};
% 		\label{fig:benefit_IC} 
        % \subcaption{(a) Resnet}
	\end{tikzpicture}
        }
        % \captionsetup{justification= centering, singlelinecheck = false}
        % \vspace{-6mm}
        \caption{Top-1 accuracy of Vanilla HDC for hyper-d from 200 to 10k (x-axis logscale)}
        
        \label{fig:acc-v-hyper-d}
        % \vspace{-6mm}s
\end{wrapfigure}
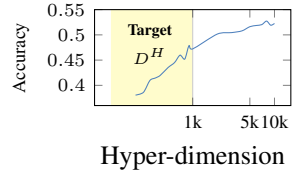

where $\delta(.)$ denotes similarity function, and for real-valued HVs it is usually cosine similarity. For bipolar HVs, this reduces to Hamming distance, where class with lowest distance is predicted.
Traditional HDC classifiers project input data into a hyper-d $>10,000$~\cite{kanerva2009-hdc}, causing overhead that is impractical for extreme edge devices . Prior works address this using sparsity~\cite{Imani2019-sparsehd}, compression~\cite{morris2019-comphd}, and quantization~\cite{imani2019-quanthd, duan2022-lehdc}, while others convert pretrained neural networks to HDC models~\cite{ma2024-hd-v-nn}, which we find suboptimal. Using smaller hyper-dimensions (e.g., $<1000$) reduces memory footprint, but also reduces accuracy sharply, as shown in Figure~\ref{fig:acc-v-hyper-d}. Figure~\ref{fig:acc-v-hyper-d} plots the top-1 accuracy of a Vanilla HDC classifier on the task of identifying 19 different types of corruptions in CIFAR10 images for hyper-d values from 200 to 10K. 
% Traditionally, HDC classifiers project input data into a hyper-d$>10,000$~\cite{kanerva2009-hdc}, which leads to significant overhead even with binary vectors, making it impractical for the extreme edge (see Appendix~\ref{apdx:hdc-in-low-hyperd}). Prior works mitigate these challenges through techniques such as dimension/class-level sparsity~\cite{Imani2019-sparsehd}, compression~\cite{morris2019-comphd} and quantization~\cite{imani2019-quanthd, duan2022-lehdc}. Other works~\cite{ma2024-hd-v-nn} simply convert a pretrained neural network to a HDC model, but we find it suboptimal. Another solution is to use smaller hyper-d values i.e., $<1000$. However, this leads to a sharp drop in accuracy. This phenomenon is illustrated in Figure~\ref{fig:acc-v-hyper-d}. Figure~\ref{fig:acc-v-hyper-d} plots the top-1 accuracy of a Vanilla HDC classifier on the task of identifying 19 different types of corruptions in CIFAR10 images for hyper-d values from 200-10K. As shown, Vanilla-HDC performs poorly in lower for hyper-d$<1000$, primarily due to suboptimal encoding. We find that Vanilla HDC, like most prior works uses a randomly-generated projection matrix because random hypervectors are nearly orthogonal to each other in a sufficiently large hyper-dimension~\cite{kanerva1998-hdc-encoding}. However, this property diminishes for hyper-d$<1000$, leading to a significant drop in performance.
We find that Vanilla HDC, like most prior works, uses a randomly-generated projection matrix since random HVs are nearly orthogonal in large hyper-dimensions~\cite{kanerva1998-hdc-encoding}. However, this property weakens for hyper-d$<1000$, causing a significant drop in accuracy.
% SparseHD~\cite{Imani2019-sparsehd} prunes the least effective elements of the trained hypervectors whereas CompHD~\cite{morris2019-comphd} compresses the class hypervectors (HV). Recent works like QuantHD~\cite{imani2019-quanthd}, LeHDC~\cite{duan2022-lehdc} quantizes the HVs from floating-point to binary/ternary models while maintaining high accuracy. 
Some prior works improve representation quality by using randomly-generated real-valued projection matrices instead of binary ones~\cite{wang2023-disthd, hernandez2021-onlinehd, zou2021-neuralhd}. While this improves accuracy, it greatly increases memory.

\section{\hddebug}
\label{sec:design}
We aim to develop a diagnostic instrument to identify the source/type of input corruption, which is highly challenging~\cite{hsu2020-generalized-odin}. 
% Although this task is complex, it minimizes the trace buffer size for logging debug information. 
Our preliminary evaluations show that a simple multi-layer perceptron (MLP) can effectively identify various corruptions but incurs significant overhead; for instance, an 8-bit 2-layer MLP for detecting 19 types of CIFAR10 corruptions uses nearly twice the resources of the ResNet-8~\cite{resnetv1} base network. Thus, we explore a HDC classifier for this task due to its lower memory/compute footprint, enabled by its bipolar nature and simple parallel operations. Despite the limitations of hyper-d $< 1000$, we aim to enhance performance through our proposed approach.
% we find a HDC classifier achieves lower memory and compute footprint owing to its bipolar nature, and simple highly parallel operations. 

\subsection{MLP-assisted Encoding}
To address the limitations of HDC in hyper-d $<1000$, we design a novel initialization scheme for the \textit{HDC encoder} where we learn the encoding/projection matrix from a simple 2-layer MLP. We observe that a 2-layer MLP (hidden+classification layer) with no bias vector closely resembles the HDC classifier. In particular, the hidden layer \emph{encodes/projects} the inputs into a higher dimension using a matrix-vector multiplication operation, which is similar to the encoding method of a HDC classifier. Figure~\ref{fig:hddebug} illustrates our proposed approach. 

\vspace{-1mm}
\textit{Training and Inference}:~As shown, we first train a 2-layer MLP model with the input dataset. 
% The input dataset is created by passing various corrupted versions of ID through the base network and recording the output of an intermediate layer. 
The hidden layer size is set to the hyper-dimension value of the HDC classifier to match HDC encoder dimensions (\tikz[baseline=(char.base)] \node[shape=circle, fill=black, text=white, inner sep=0.5pt] (char) {A};). The weights learned by the hidden layer of the MLP with dimensions $d\times D^H$ is fed to the \emph{encoder} module after passing through the $sgn(.)$ function (\tikz[baseline=(char.base)] \node[shape=circle, fill=black, text=white, inner sep=0.5pt] (char) {B};). Next, the encoder module uses the \textit{learnt projection matrix} to convert all feature vectors in the train set to hypervectors ($\mathcal{H}s$) with hyper-dimension $D^H$ (\tikz[baseline=(char.base)] \node[shape=circle, fill=black, text=white, inner sep=0.5pt] (char) {C};). Finally, the training module computes all the class HVs (\tikz[baseline=(char.base)] \node[shape=circle, fill=black, text=white, inner sep=0.5pt] (char) {D};), and performs inference as described in Section~\ref{subsec:hdc_prelim}. The MLP's projection matrix allows HDC's encoder to better separate data in the hyperspace, leading to a significant improvement in its performance.

\section{Results and Discussion}
\label{sec:results}
% NOTES
% - add accuracy of all HDC . In the smae table make two bifurcations, binary and non-binary
% - explain why top-2 and top3 accuracy are good enough for debug and show commonly occuring top-3 corruption candidates.
% - in the appendix, add SSIM tables for all datasets
% \begin{wrapfigure}{r}{0.7\textwidth}  
We evaluate \hddebug on one audio and two image classification tasks. We use the in-distribution (ID) and corrupted-in-distribution (CID) versions of 1)~SpeechCmd~\cite{speech_commands} on a 4-layer depthwise-separable model (DSCNN), 2)~TinyImagnet~\cite{le2015-tinyimagenet} on a MobilenetV2~\cite{mobilenets} and 3)~CIFAR10~\cite{cifar-10} on Resnet-8 from MLPerf Tiny benchmarck suite~\cite{mlperf-tiny-benchmark}. For CID datasets, we use CIFAR10-C (19 corruptions) and TinyImagenet-C (15 corruptions) from~\cite{hendrycks2019-benchmarking-corruptions}. For SpeechCmd, we use the audiomentations library~\cite{jordal2024-audiomentations} to obtain SpeechCmd-C with 11 corruptions. To avoid the extra costs of on-device preprocessing and feature engineering, we directly utilize the output of an intermediate layer of the base network as input to HDC. However, this output may not be the most informative, which limits overall performance of the HDC classifier. Appendix~\ref{apdx:dataset} has more details.

\vspace{-1mm}
\textbf{Evaluation and metrics}~We evaluate the accuracy of all HDC baselines on the task of detecting corruptions. In our evaluations with prior HDC works, we differentiate them into binary~\cite{duan2022-lehdc, imani2019-adapthd, morris2019-comphd, imani2019-quanthd, Imani2019-sparsehd} and non-binary~\cite{wang2023-disthd, zou2021-neuralhd, hernandez2021-onlinehd} HDC. We use the torch-hd library~\cite{heddes2023-torchhd} to implement prior works. We envision a scenario where a finite set of \textit{commonly-occurring} corruptions~\cite{hendrycks2019-benchmarking-corruptions} affecting inputs are used to train the HDC before deployment. Post-deployment, the HDC classifier is invoked only when the model monitoring mechanism detects an accuracy drop in the base network to identify corruption causing the drop. To balance resource usage and accuracy, we use hyper-d values of 300, 400 and 300 for CIFAR10, T-imgnet and SpCmd respectively. See Appendix~\ref{apdx:eval} for more details.
\begin{figure}
    \centering
    \captionsetup{justification= centering, singlelinecheck = false}
    \hspace{-1cm}
    \begin{subfigure}[]{0.3\textwidth}
        % \renewcommand{\arraystretch}{1.5} % Adjust row height (increase this value if needed)
% \begin{table}[t]
% \centering
% \small
\scalebox{0.65}{

{

% \begin{tabular}{|| c | c | c | c | c | c | c | c | c | c | c ||}
\begin{tabular}{l | c | c | c | c  }
 & \textbf{Baseline}&  \textbf{CIFAR10}  & \textbf{T-imgnet}  & \textbf{SpCmd}   \\
\hline
 \multirow{7}{*}{\shortstack{\scriptsize B\\\scriptsize I\\\scriptsize N\\\scriptsize A\\\scriptsize R\\\scriptsize Y}} & Vanilla & 0.42 & 0.71 & 0.35\\

 & LeHDC & 0.29 & 0.64 & 0.22\\
 & AdaptHD & 0.34 & 0.75 & 0.32\\

 & CompHD & 0.45 & 0.69 & 0.32\\

 & QuantHD & 0.38 & 0.65 & 0.23\\

 & SparseHD & 0.34 & 0.65 & 0.25\\

 & \textbf{\hddebug} & \textbf{0.68} & \textbf{0.78} & \textbf{0.45}\\

\hline
\hline

\multirow{3}{*}{\shortstack{\scriptsize R\\\scriptsize E\\\scriptsize A\\\scriptsize L}} & DistHD & \textbf{0.78} & 0.84 & 0.43\\

& NeuralHD & 0.74 & 0.88 & \textbf{0.51}\\

& OnlineHD & 0.75 & \textbf{0.89} & 0.50\\

\hline
\hline
& MLP & 0.85 & 0.90 & 0.54\\

\end{tabular}
}
}

% \caption{Top-1 Accuracy of HDC classifiers on task of identifying corruptions. Bold marks best results}
% \label{table:cid-detect-mnist}
% \end{table}
    \caption{Top-1 Accuracy}
    \label{fig:table-acc}
    \end{subfigure}
    \hspace{1.7cm}
    \begin{subfigure}[]{0.28\textwidth}
            \centering
    \scalebox{0.9}{
        \begin{tikzpicture}
	\pgfplotstableread{
        x 	img-size	hdc-size img-dim base-size
        % 1 	1	10 mnst 4
        2 	3	10 cfr 145.91
        4 	12	10 timgnet 2580
        6 	3.92 5.2 spcmd 78.2
        }\data
		% \begin{axis}
            \begin{semilogyaxis}
                [
			% Chart dimensions
			width=\textwidth,
			height=\graphHeight*2 - 3.5cm,
			%
			% Ticks config
			% Removing smaller tick lines since grids are used
			major tick length=3pt,
			%
			% Grid lines config
			% Y axis grids are major grids and X axis grids are minior grids
			major grid style={dashed,color=gray!50},
			minor grid style={color=gray!50},
			% ymajorgrids=true,
			% xmajorgrids=true,
% 			xminorgrids=true,
			%
			% X axis config
% 			xtick=data,
% 			xticklabels from table={\data}{ee-placement},
                % symbolic x coords={28x28x1, 32x32x3, 64x64x3, 96x96x3},
                % symbolic x coords={mnst, cfr, timgnet, vww},
			xticklabel style={text height=5pt,font=\scriptsize},
			%
			% X label config
			% xticklabels from table={\data}{img-dim},
                xticklabels={,, cfr, T-img, spcmd} ,  
			xlabel style={font=\scriptsize},
		    xlabel near ticks,
                xmin = 2,
                xmax =6,
                scaled x ticks=false,
                % xtick = {1,2,3,4},
                % extra x ticks={0, 500, 1000, 5000,10000},
                % xticklabels={ 1k, 5k, 10k},
		      % extra x ticks={0,200, 500, 700, 1000, 5000,10000},
			%extra x tick style={
			%	grid=minor,
		    %	xticklabel=\empty
			%},
			%
			% Y axis config
			ymin=-1,ymax=3000,
			yticklabel style={font=\scriptsize},
			%
			% Y label config
			ylabel style={align=center,font=\scriptsize},
			ylabel=Size (KB),
			% ylabel near ticks,
			% Legend config
			legend style={at={(0.55,1.0)},anchor=south,cells={align=left},draw=none,font=\scriptsize,text width=1cm},
			legend columns=-1,
                % legend pos= north ,
			%
			% Bar config
			ybar=3pt,
			area legend,
			bar width=3pt,
			enlarge x limits=0.3,			
			%
			% Text on each bar config
			% point meta set to explicit since the log scale changes the value of \pgfplotspointsmeta
			%point meta=explicit symbolic,
			%nodes near coords={\textbf{{\pgfplotspointmeta}}},
			%every node near coord/.append style={font=\tiny, rotate=90, anchor=west},
			%nodes near coords align={vertical}
		]
             % Highlight the region from x=0 to x=1000
            % \addplot[fill=yellow, opacity=0.2] coordinates {(0,0.36) (1000,0.36) (1000,0.55) (0,0.55)};
                % Annotation for the highlighted region
            % Highlight the region from x=0 to x=1000
            % \addplot+[fill=yellow, opacity=0.2] coordinates {(100,0.36) (1000,0.36) (1000,0.55) (100,0.55)};
            % % Annotation for the highlighted region
            % \node at (axis cs: 300,0.54) [align=center, anchor=north] {\tiny \textbf{Target} \\ \tiny \textbf{$D^H$}};

		% \addplot [bar1] table[x=x, y=img-size, ] {\data};
            \addplot [bar2] table[x=x, y=hdc-size, ] {\data};
            \addplot [bar3] table[x=x, y=base-size, ] {\data};

% 		\addplot +[mark=none] coordinates {(2, -1) (2, 3)};
% 		\addplot [ppurple, mark=*] table[x=ee-placement,y=ee-accuracy] {\data};
% 		\addplot [rred] table[x=X,y=mkr-speed] {\maxscaleData};
		
		%\addplot [bar1] table[x=X,y=uno-speed] {\maxscaleData};
		%\addplot [bar2] table[x=X,y=mkr-speed] {\maxscaleData};
		
		% \addlegendentry{Img size}
            \addlegendentry{\tiny HDC}
            \addlegendentry{\tiny base model}
  %             \addlegendentry{3stack}
  %             \addlegendentry{4stack}

% 		\addlegendentry{Reference}

        % \node (resnet) at (anchor.south) [ionode,black,  xshift=0.5cm, yshift=-1.5cm] {a) Resnet};
        
		\end{semilogyaxis}
            % \end{axis}
  
% 		\node (anchor) at (0,0) [ionode,black, xshift=2.7cm, yshift=-5.5] {a) Resnet};
% 		\label{fig:benefit_IC} 
        % \subcaption{(a) Resnet}
	\end{tikzpicture}
    }
    \caption{Base model and \hddebug size comparison. (y-axis logscale)}
    \label{fig:size}
    \end{subfigure}
    \hspace{0.0cm}
    \begin{subfigure}[]{0.2\textwidth}
            \centering
    \scalebox{0.8}{
        % \renewcommand{\arraystretch}{1.5} % Adjust row height (increase this value if needed)
% \begin{table}[t]
% \centering
% \small
\scalebox{1}{

{

% \begin{tabular}{|| c | c | c | c | c | c | c | c | c | c | c ||}
\begin{tabular}{l | c | c  }
 \textbf{Dataset}&  \textbf{Top-2} & \textbf{Top-3} \\
\hline
 CIFAR10 & 0.86 & 0.95\\

 T-imgnet & 0.92 & 0.95\\

 SpCmd & 0.58 & 0.67\\

\end{tabular}
}
}

% \caption{Top-1 Accuracy of HDC classifiers on task of identifying corruptions. Bold marks best results}
% \label{table:cid-detect-mnist}
% \end{table}
    }
    \caption{Top-2 and Top-3 accuracy of \hddebug}
    \label{fig:table-topk}
    \end{subfigure}
    \caption{Experimental results}
    \label{fig:results}
    \vspace{-3mm}
\end{figure}
% \end{wrapfigure}

\vspace{-1mm}
\textbf{HDC Results}:~Table~\ref{fig:table-acc} reports the top-1 accuracy of all baselines. \hddebug outperforms all binary HDC methods across evaluated datasets, achieving an average accuracy \textbf{improvement of 12\%} over the best-performing binary HDC work, highlighting the effectiveness of the MLP-assisted encoding scheme.
Surprisingly, Vanilla often surpasses many state-of-the-art binary HDC works. Further analyses reveal that unlike Vanilla and \hddebug's single-pass training, binary baselines employ a retraining step over multiple epochs, with some methods~\cite{duan2022-lehdc, imani2019-quanthd} regenerating the least contributing hyper-dimensions by changing the encoding. However, we find that with lower hyper-d values (e.g., 300, 400), changes to encoder after a train-pass leads to an immediate drop in accuracy, with the next epoch barely recovering the lost performance, resulting in suboptimal outcomes. In contrast, higher hyper-d values allow for less accuracy drop between epochs during regeneration. For instance, accuracies of both Vanilla and AdaptHD on SpCmd for hyper-d=1000 is 0.39. Additionally, most binary HDC methods maintain two versions of class HV: non-binary for training updates and binary for inference/validation, typically obtained by applying the $sgn(.)$ function. This approach also causes sharp accuracy drops at lower hyper-d values.
We also report the accuracy of a conventional MLP as a reference. While non-binary HDC approaches are close to MLP in performance due to superior data representation, binary HDC methods lag behind. However, \hddebug still achieves strong performance relative to the MLP.

% As shown, \hddebug outperforms all binary HDC works on all evaluated datasets we evaluate. In many cases, Vanilla outperforms state-of-the-art prior works in binary HDC. Further analyses showed that unlike Vanilla which uses a single-pass training, all binary baselines implement a retraining step that fine-tunes the HDC model over several iterations (epochs) of training. Some works~\cite{duan2022-lehdc, imani2019-quanthd} determine the lowest contributing hyper-dimensions in the hyperspace and \textit{regenerate} those dimensions by changing the encoding. However, we find that when working with lower hyper-d values (e.g., 300, 400), changing the encoder results in an immediate drop in accuracy. After changing the encoder to regenerate dimensions, the next epoch barely manages to recover the lost accuracy, resulting in suboptimal performance. In contrast, with higher hyper-d values dimension regeneration causes lesser drop in accuracy between epochs, leading to better performance. For example, for hyper-d=1000, top-1 accuracy of Vanilla is  and that of LeHDC~\cite{duan2022-lehdc} is 0.42. In addition, most binary HDC works maintain two versions of class HVs: non-binary to update during training and binary for inference/validation. The latter is usually obtained by passing the former through $sgn(.)$. This too results in an immediate drop in accuracy due lower hyper-d values. 

\vspace{-1mm}
\textbf{Debugging}~With multiple \textit{similar} corruptions (see Table~\ref{table:ssim-spcmd} and Table~\ref{table:ssim-cifar10}) and inherent model limitations, HDC's top-1 accuracy is constrained. For SpCmd, we observe that \textit{highly similar} corruptions (Table~\ref{table:ssim-spcmd}) and the quality of the intermediate layer data limits \hddebug's performance, impacting even the MLP. This reflects an intrinsic limitation of the data, which could potentially be addressed through improved data or feature engineering. Table~\ref{fig:table-topk} presents the top-2 and top-3 accuracy of \hddebug, demonstrating that on-device diagnosis can effectively narrow down corruptions to the top candidates with high precision. This advancement is vital as it improves the efficiency of diagnosing model failures by reducing the number of potential error sources for developers to investigate, particularly when faced with multiple corruption types. Figure~\ref{fig:size} illustrates the overhead of \hddebug relative to the base network. We observe that the trade-off between accuracy and overhead improves with larger model sizes and higher-resolution images, both of which are realistic scenarios in practical applications. While our work focuses on identifying image corruptions—a specific subset of potential model failures—there are numerous other diagnostics that can be applied on-device across a wide range of applications~\cite{ochiai2019-debug-use, ren2021-tinyol, lin2022-ondeviceml-mcu}. This work serves as a strong foundation to facilitate the exploration.

\section{Conclusion}
\label{sec:conclusion}
We present \hddebug, a resource-efficient approach for on-device debugging in TinyML devices using HDC. \hddebug leverages an improved HDC encoding technique, learned through neural networks, to outperform previous binary HDC methods in identifying corruptions in inputs across various datasets. With HDC’s inherent fault tolerance~\cite{liu2019-hdc-fault-tolerance} alongside the proposed enhanced encoding, \hddebug strengthens TinyML reliability. This work underscores HDC’s potential as a robust foundation for integrating diagnostics in TinyML, facilitating advancements in model reliability and performance through innovative, real-time, on-device debugging solutions.
% offers significant improvements in both accuracy and efficiency, making it a valuable tool for ensuring reliability in resource-constrained environments. This work highlights HDC as a solid foundation for integrating debugging tools in TinyML, driving advancements in model reliability through innovative diagnostics.

\vspace{-1mm}
\textbf{Limitations}:~HDC methods are generally outperformed by conventional neural networks (NNs) and often require specialized parallel hardware, such as FPGAs, to fully leverage their computational efficiency~\cite{redding2023-embhd}. Their performance is also influenced by the number of corruptions to identify; however, it can improve if developers narrow the focus to specific corruptions (see Appendix~\ref{apdx:debug}). Additionally, the accuracy gain from MLP-assisted encoding is notable only in smaller hyperspaces. For instance, the accuracies of AdaptHD and \hddebug are 0.83 and 0.84, respectively, for hyper-d=2000 on TinyImageNet. Finally, we observe that incorporating \hddebug into extremely small models on toy datasets (e.g., Ward~\cite{ward}, USPS~\cite{usps}, MNIST~\cite{mnist}) incurs significant overhead relative to the base model size. Nevertheless, \hddebug operates efficiently with more realistic models.

\bibliographystyle{ACM-Reference-Format}
\bibliography{bibliography}

%%%%%%%%%%%%%%%%%%%%%%%%%%%%%%%%%%%%%%%%%%%%%%%%%%%%%%%%%%%%
\newpage
\appendix

\section{Appendix}

\subsection{Dataset details}
\label{apdx:dataset}
In this section, we describe the details of the datasets and models evaluated.

\subsubsection{Datasets}
In our evaluations, we use three in-distribution datasets for training all baselines methods we evaluate: 1)~SpeechCmd~\cite{speech_commands}, 2)~CIFAR10~\cite{cifar-10} and 3)~TinyImagenet~\cite{le2015-tinyimagenet}. 

% \textbf{MNIST}~
% MNIST is a dataset of handwritten digits containing 60,000 grayscale images of size 28$\times$28 and a test set of 10,000 images. MNIST contains 10 classes.

\textbf{SpeechCommands}~
Speech-Cmd is a collection of short audio clips, each spanning 1 second. The dataset consists of utterances for 35 words and is commonly used for benchmarking keyword spotting systems. We train our model (DSCNN) to recognize ten words out of 35: \textit{Down, Go, Left, No, Off, On, Right, Stop, Up, Yes}. Thus, the number of classes is 10. The audio files in WAV format are preprocessed to compute Mel-frequency cepstral coefficients (mel-spectograms). The mel-spectograms are of size 49$\times$10 with a single channel. 
% The DSCNN model is trained for 10 epochs with a batch size of 100.

\textbf{CIFAR10}~
CIFAR10 dataset consists of 60,000 32$\times$32 rgb images out of which 10,000 images are in the test set. It contains 10 classes and thus 6000 images per class.

\textbf{TinyImagenet}~
TinyImagenet is a smaller version of the Imagenet~\cite{imagenet} dataset containing 200 classes instead of 1000 classes of the original Imagenet. Each class in TinyImagenet has 500 images in the train set and the validation set contains 50 images per class. The size of the images are resized and fixed at 64$\times$64$\times$3.

\subsubsection{Corrupted-in-distribution (CID) datasets}
We use the following corrupted versions of ID: 1)~CIFAR10-C and 2)~TinyImagenet-C from~\cite{hendrycks2019-benchmarking-corruptions}. For both, we use the corruptions with severity level=5 in our evaluations. All corruptions are drawn from  4 major sources: noise, blur, weather and digital.
We create SpeechCmd-C from the audiomentations libary~\cite{jordal2024-audiomentations}.

% \textbf{MNIST-C}~
% The MNIST-C dataset contains 15 corrupted versions of MNIST - \textit{shot\_noise, impulse\_noise, glass\_blur, fog, spatter, dotted\_line, zigzag, canny\_edges, motion\_blur, shear, scale, rotate, brightness, translate, stripe, identity}. All the corruptions are of a fixed severity level. 

\textbf{CIFAR10-C}~
CIFAR10-C includes 19 different types of corruptions with 5 severity levels. The list of corruptions are: \textit{gaussian\_noise, brightness, contrast, defocus\_blur, elastic, fog, frost, frosted\_glass\_blur, gaussian\_blur,  impulse\_noise,  jpeg\_compression, motion\_blur, pixelate, saturate, shot\_noise,  snow, spatter, speckle\_noise and zoom\_blur}. 

\textbf{TinyImagenet-C}~
TinyImagenet-C includes 15 different types of corruptions with 5 severity levels. The list of corruptions are: \textit{gaussian\_noise, brightness, contrast, defocus\_blur, elastic\_transform, fog, frost, glass\_blur, impulse\_noise,  jpeg\_compression, motion\_blur, pixelate, shot\_noise,  snow and zoom\_blur}. 

\textbf{SpeechCmd-C}~For keyword spotting (KWS)~\cite{hello-edge}, an audio classification task on the Speech Commands dataset, we introduce noise and other corruptions to the audio using the audiomentations library~\cite{jordal2024-audiomentations}. We apply the following 11 corruptions: \textit{gaussian noise, air absorption, band pass filter, band stop filter, high pass filter, high shelf filter, low pass filter, low shelf filter, peaking filter, tanh distortion, time mask and time stretch}.

\subsection{Experimental Setup}
We evaluate three models in our experiments:1)~4-layer DSCNN on SpeechCmd~\cite{hello-edge}, 2)~Resnet-8 with 3 residual stacks from the MLPerf Tiny benchmark suite~\cite{mlperf-tiny-benchmark} on CIFAR10 and 3)~MobilenetV2~\cite{mobilenets} on TinyImagenet.

The 4-layer DSCNN is trained for 10 epochs with a batch size of 100, the Resnet-8 model is trained for 200 epochs with batch size of 32 and the MobilenetV2 model is trained for 200 epochs with a batch size of 128. All models are trained with Adam optimizer with momentum of 0.9 and an initial learning rate of 0.001, expect DSCNN on SpeechCmd which uses an initial learning rate of 0.0005. The learning rate is decayed by a factor of 0.99 every epoch for image classification datasets, and we follow a step function for SpeechCmd that reduces learning rate by half every 2 epochs. 

We train the MLP used to learn the encoding matrix for 20 epochs with a batch size of 256.

\subsection{Dataset creation for HDC classifier}
\label{apdx:dataset_hdc}
Since the HDC classifier aims to distinguish between different types of corruptions in the input images, the natural input to the classifier would be the images themselves. However, the high dimensionality of the input images (e.g., 12K for TinyImageNet) significantly increases the resource usage of the HDC classifier. One potential solution is to reduce the dimensionality of the input images, but this introduces additional processing costs that may impact the execution of the base network.

To avoid such preprocessing or feature engineering overhead, we propose tapping into an appropriate intermediate layer of the base model directly. This approach offers two key advantages: 1)~as the input image passes through the base network, its dimensionality is reduced, which in turn minimizes the size of the HDC classifier, and 2)~ the execution of the base network remains uninterrupted—an important consideration for safety-critical applications.

The dataset for the HDC classifier is created as follows.

1)~First, we iterate through all the corrupted-ID images for each dataset by passing each corrupted image through the pretrained base network. The base network has been trained only on ID.

2)~Second, we select an appropriate intermediate layer of the base network, and record the output of this intermediate layer as we iterate through the corrupted images. We use this to create our train and test datasets for the HDC classifier denoted by $\mathcal{D}_{HDC-train}$ and $\mathcal{D}_{HDC-test}$ respectively.

3)~Next, we use $\mathcal{D}_{HDC-train}$ to train the HDC classifier as described in Section~\ref{sec:design}, and use $\mathcal{D}_{HDC-test}$ for validation/testing.

\subsubsection{Selecting an appropriate intermediate layer}

Since tinyML models often have only a few layers, the number of intermediate layers available for tapping is limited. Therefore, we perform an exhaustive search. We construct $\mathcal{D}_{HDC-train}$ and $\mathcal{D}_{HDC-test}$ as described above by taping into each intermediate layer's output. Next, we train and test a Vanilla HDC using each version of the dataset. We select the intermediate layer whose dataset yields the highest validation accuracy.
For SpCmd using a 4-layer DSCNN, the tapping location is after the first depthwise layer. For CIFAR10 using a 3-residual stack Resnet, the best location is after the third residual stack. In the case of TinyImageNet, which employs a larger MobileNetV2 model with 17 inverted residual blocks, we refine our search using a binary search approach. This systematic method helps us narrow down the most informative layer by evaluating performance at progressively finer intervals. For MobileNetV2, we find that the output of the 5\textsuperscript{th} inverted residual block is the most informative. The number of features in the HDC datasets thus obtained are 256, 128 and 192 for CIFAR10, SpCmd and T-imgnet respectively.

\subsection{Evaluation details}
\label{apdx:eval}
We distinguish the HDC baselines we compare against into two types: 1)~Binary HDC methods:~LeHDC~\cite{duan2022-lehdc}, AdaptHD~\cite{imani2019-adapthd}, CompHD~\cite{morris2019-comphd}, QuantHD~\cite{imani2019-quanthd} and SparseHD~\cite{Imani2019-sparsehd}, and 2)~Non-binary HDC methods:~DistHD~\cite{wang2023-disthd}, NeuralHD~\cite{zou2021-neuralhd} and OnlineHD~\cite{hernandez2021-onlinehd}. 

\textbf{Simulating model accuracy drop}~In our evaluations, we trigger the proposed debug tool only when the model monitoring mechanism detects an accuracy drop.~\citet{ghanathe2024-qute} proposes a resource-efficient accuracy monitoring mechanism for tinyML models with minimal resource overhead. We follow the procedure outlined in~\cite{ghanathe2024-qute} to simulate the accuracy drop detection. First, we use the base network to iterate over all in-distribution (ID) samples using a sliding window of size 100, calculating accuracy over the past 100 input samples. The resulting accuracy distribution is denoted as $\mathcal{A}_{ID}$, with mean $\mu_{ID}$ and standard deviation $\sigma_{ID}$. Next, we create ID+CID datasets by appending ID with all corrupted-ID versions. We then iterate over the ID+CID datasets using the base network and the same sliding window approach. The HDC classifier is only invoked when $\mathcal{A}_{SW} < \mu_{ID} - 3\cdot \sigma_{ID}$, where $\mathcal{A}_{SW}$ is accuracy of sliding window. Once invoked, the HDC classifier reads the output of the predetermined intermediate layer and identifies the type of corruption that caused the model accuracy to drop.

\subsubsection{MLP-assisted encoding}
A HDC classifier for debugging in tinyML devices necessitates maintaining the bipolar nature of the network while using hyper-dimensional values of less than 500 to ensure that the associated overhead remains practical.
However, we observe that in a lower hyper-dimensional space (few hundreds of dimensions), the random projection matrix is inadequate in separating the data in the hyperspace, which in turn significantly limits the performance of the HDC model. This phenomenon is illustrated in Figure~\ref{fig:acc-v-hyper-d}. In our initial experiments, we observe two things: 1)~a simple 2-layer MLP has a close resemblance to a HDC model and 2)~the MLP is able to achieve significantly higher accuracy compared to its HDC counterpart. This demonstrates that the single hidden layer of the MLP is able to project data into the hyperspace much more efficiently than the random-projection matrix. Since the hidden layer functions similar to the HDC encoder, we import the weight matrix into the encoder module of the HDC classifier, which proves to be crucial in extracting maximum performance from the HDC classifier, especially when working with low hyper-d values ($<1000$).

We made several modifications to the MLP architecture used for learning the encoding matrix. In one iteration, we included a bias vector followed by an activation function. However, we found that the encoding matrix was negatively impacted by the additional assistance from the bias vector and activation function. As a result, we adopted a configuration where the hidden layer output has neither a bias vector nor an activation function. 

% We also compare \hddebug with~\citet{ma2024-hd-v-nn}, which derives 

\definecolor{yellow}{rgb}{1.0,1.0,0.6}
\definecolor{green}{rgb}{0.6,1.0,0.6}

\begin{table}[ht]
\centering
\small
\scalebox{0.7}{
\begin{tabular}{l | c | c | c | c | c | c | c | c | c | c | c | c | c | c | c | c | c | c | c | c}
\textbf{SSIM} & \textbf{ID} & \textbf{GN} & \textbf{BR} & \textbf{CT} & \textbf{DB} & \textbf{EL} & \textbf{FG} & \textbf{FR} & \textbf{FGB} & \textbf{GB} & \textbf{IN} & \textbf{JC} & \textbf{MB} & \textbf{PX} & \textbf{ST} & \textbf{SN} & \textbf{SW} & \textbf{SP} & \textbf{SpN} & \textbf{ZB} \\
\hline
\textbf{ID} & 	\cellcolor{green}1.00  &	0.46  &	\cellcolor{green}0.81  &	0.34  &	\cellcolor{green}0.76  &	\cellcolor{yellow}0.64  &	\cellcolor{yellow}0.56  &	\cellcolor{yellow}0.68  &	0.46  &	\cellcolor{green}0.75  &	\cellcolor{yellow}0.65  &	\cellcolor{green}0.80  &	\cellcolor{yellow}0.64  &	\cellcolor{green}0.78  &	\cellcolor{yellow}0.70  &	\cellcolor{yellow}0.50  &	\cellcolor{yellow}0.71  &	\cellcolor{green}0.83  &	\cellcolor{yellow}0.53  &	\cellcolor{yellow}0.61\\  \hline

\textbf{GN} & 	0.46  &	\cellcolor{green}1.00  &	0.35  &	0.10  &	0.34  &	0.25  &	0.25  &	0.34  &	0.19  &	0.29  &	0.32  &	0.40  &	0.25  &	0.35  &	0.35  &	0.32  &	0.37  &	0.41  &	0.34  &	0.26\\  \hline

\textbf{BR} & 	\cellcolor{green}0.83  &	0.37  &	\cellcolor{green}1.00  &	0.34  &	\cellcolor{yellow}0.61  &	0.50  &	\cellcolor{yellow}0.50  &	\cellcolor{yellow}0.64  &	0.40  &	\cellcolor{yellow}0.61  &	\cellcolor{yellow}0.52  &	\cellcolor{yellow}0.62  &	\cellcolor{yellow}0.54  &	\cellcolor{yellow}0.64  &	\cellcolor{yellow}0.54  &	0.38  &	\cellcolor{yellow}0.70  &	\cellcolor{yellow}0.65  &	0.39  &	\cellcolor{yellow}0.51\\  \hline

\textbf{CT} & 	0.45  &	0.15  &	0.45  &	\cellcolor{green}1.00  &	\cellcolor{yellow}0.56  &	0.45  &	\cellcolor{yellow}0.57  &	0.38  &	0.37  &	\cellcolor{yellow}0.60  &	0.30  &	0.41  &	0.50  &	0.46  &	0.38  &	0.17  &	0.36  &	0.38  &	0.19  &	0.49\\  \hline

\textbf{DB} & 	\cellcolor{green}0.75  &	0.32  &	\cellcolor{yellow}0.60  &	0.37  &	\cellcolor{green}1.00  &	\cellcolor{yellow}0.72  &	\cellcolor{yellow}0.53  &	\cellcolor{yellow}0.55  &	\cellcolor{yellow}0.54  &	\cellcolor{green}0.98  &	0.50  &	\cellcolor{green}0.75  &	\cellcolor{yellow}0.71  &	\cellcolor{green}0.78  &	\cellcolor{yellow}0.55  &	0.36  &	\cellcolor{yellow}0.52  &	\cellcolor{yellow}0.65  &	0.38  &	\cellcolor{yellow}0.73\\  \hline

\textbf{EL} & 	\cellcolor{yellow}0.63  &	0.29  &	\cellcolor{yellow}0.52  &	0.26  &	\cellcolor{yellow}0.72  &	\cellcolor{green}1.00  &	0.44  &	0.44  &	0.46  &	\cellcolor{yellow}0.72  &	0.41  &	\cellcolor{yellow}0.59  &	\cellcolor{yellow}0.57  &	\cellcolor{yellow}0.62  &	0.47  &	0.32  &	0.43  &	\cellcolor{yellow}0.53  &	0.34  &	\cellcolor{yellow}0.57\\  \hline

\textbf{FG} & 	\cellcolor{yellow}0.57  &	0.27  &	\cellcolor{yellow}0.51  &	0.38  &	\cellcolor{yellow}0.56  &	0.46  &	\cellcolor{green}1.00  &	0.46  &	0.33  &	\cellcolor{yellow}0.56  &	0.39  &	\cellcolor{yellow}0.50  &	0.48  &	0.50  &	0.44  &	0.29  &	0.43  &	0.49  &	0.30  &	0.47\\  \hline

\textbf{FR} & 	\cellcolor{yellow}0.69  &	0.35  &	\cellcolor{yellow}0.68  &	0.26  &	\cellcolor{yellow}0.56  &	0.44  &	0.50  &	\cellcolor{green}1.00  &	0.31  &	\cellcolor{yellow}0.55  &	0.45  &	\cellcolor{yellow}0.59  &	0.47  &	\cellcolor{yellow}0.53  &	0.47  &	0.35  &	\cellcolor{yellow}0.57  &	\cellcolor{yellow}0.57  &	0.35  &	0.44\\  \hline

\textbf{FGB} & 	0.45  &	0.20  &	0.37  &	0.20  &	\cellcolor{yellow}0.52  &	0.45  &	0.30  &	0.34  &	\cellcolor{green}1.00  &	\cellcolor{yellow}0.53  &	0.32  &	0.43  &	0.39  &	0.48  &	0.38  &	0.23  &	0.32  &	0.41  &	0.21  &	\cellcolor{yellow}0.53\\  \hline

\textbf{GB} & 	\cellcolor{green}0.77  &	0.31  &	\cellcolor{yellow}0.62  &	0.41  &	\cellcolor{green}0.98  &	\cellcolor{yellow}0.72  &	\cellcolor{yellow}0.57  &	\cellcolor{yellow}0.54  &	\cellcolor{yellow}0.54  &	\cellcolor{green}1.00  &	0.50  &	\cellcolor{yellow}0.71  &	\cellcolor{yellow}0.74  &	\cellcolor{green}0.76  &	\cellcolor{yellow}0.55  &	0.36  &	0.50  &	\cellcolor{yellow}0.64  &	0.38  &	\cellcolor{yellow}0.72\\  \hline

\textbf{IN} & 	\cellcolor{yellow}0.64  &	0.34  &	\cellcolor{yellow}0.51  &	0.19  &	0.48  &	0.40  &	0.34  &	0.46  &	0.31  &	0.48  &	\cellcolor{green}1.00  &	\cellcolor{yellow}0.53  &	0.40  &	\cellcolor{yellow}0.51  &	0.48  &	0.38  &	0.45  &	\cellcolor{yellow}0.55  &	0.35  &	0.38\\  \hline

\textbf{JC} & 	\cellcolor{green}0.79  &	0.41  &	\cellcolor{yellow}0.60  &	0.25  &	\cellcolor{yellow}0.71  &	\cellcolor{yellow}0.58  &	0.47  &	\cellcolor{yellow}0.57  &	0.44  &	\cellcolor{yellow}0.71  &	\cellcolor{yellow}0.54  &	\cellcolor{green}1.00  &	\cellcolor{yellow}0.60  &	\cellcolor{yellow}0.70  &	\cellcolor{yellow}0.60  &	0.41  &	\cellcolor{yellow}0.59  &	\cellcolor{yellow}0.69  &	0.46  &	\cellcolor{yellow}0.53\\  \hline

\textbf{MB} & 	\cellcolor{yellow}0.65  &	0.30  &	\cellcolor{yellow}0.56  &	0.36  &	\cellcolor{yellow}0.69  &	\cellcolor{yellow}0.59  &	0.48  &	0.48  &	0.40  &	\cellcolor{yellow}0.73  &	0.42  &	\cellcolor{yellow}0.60  &	\cellcolor{green}1.00  &	\cellcolor{yellow}0.63  &	0.49  &	0.29  &	0.44  &	\cellcolor{yellow}0.54  &	0.33  &	\cellcolor{yellow}0.57\\  \hline

\textbf{PX} & 	\cellcolor{green}0.76  &	0.35  &	\cellcolor{yellow}0.63  &	0.29  &	\cellcolor{green}0.77  &	\cellcolor{yellow}0.63  &	0.46  &	\cellcolor{yellow}0.54  &	0.46  &	\cellcolor{green}0.76  &	\cellcolor{yellow}0.52  &	\cellcolor{yellow}0.73  &	\cellcolor{yellow}0.60  &	\cellcolor{green}1.00  &	\cellcolor{yellow}0.60  &	0.41  &	\cellcolor{yellow}0.54  &	\cellcolor{yellow}0.68  &	0.41  &	\cellcolor{yellow}0.58\\  \hline

\textbf{ST} & 	\cellcolor{yellow}0.68  &	0.35  &	\cellcolor{yellow}0.56  &	0.24  &	\cellcolor{yellow}0.55  &	0.44  &	0.40  &	0.46  &	0.33  &	\cellcolor{yellow}0.54  &	0.49  &	\cellcolor{yellow}0.56  &	0.48  &	\cellcolor{yellow}0.58  &	\cellcolor{green}1.00  &	0.37  &	0.46  &	\cellcolor{yellow}0.60  &	0.38  &	0.45\\  \hline

\textbf{SN} & 	0.48  &	0.34  &	0.37  &	0.10  &	0.32  &	0.29  &	0.26  &	0.35  &	0.22  &	0.35  &	0.36  &	0.45  &	0.29  &	0.38  &	0.38  &	\cellcolor{green}1.00  &	0.37  &	0.42  &	0.36  &	0.26\\  \hline

\textbf{SW} & 	\cellcolor{yellow}0.72  &	0.40  &	\cellcolor{yellow}0.72  &	0.24  &	\cellcolor{yellow}0.54  &	0.43  &	0.42  &	\cellcolor{yellow}0.58  &	0.30  &	\cellcolor{yellow}0.53  &	0.45  &	\cellcolor{yellow}0.60  &	0.46  &	\cellcolor{yellow}0.53  &	0.45  &	0.40  &	\cellcolor{green}1.00  &	\cellcolor{yellow}0.58  &	0.42  &	0.40\\  \hline

\textbf{SP} & 	\cellcolor{green}0.83  &	0.40  &	\cellcolor{yellow}0.67  &	0.26  &	\cellcolor{yellow}0.62  &	\cellcolor{yellow}0.52  &	0.47  &	\cellcolor{yellow}0.58  &	0.40  &	\cellcolor{yellow}0.64  &	\cellcolor{yellow}0.56  &	\cellcolor{yellow}0.69  &	\cellcolor{yellow}0.54  &	\cellcolor{yellow}0.68  &	\cellcolor{yellow}0.57  &	0.45  &	\cellcolor{yellow}0.57  &	\cellcolor{green}1.00  &	0.46  &	\cellcolor{yellow}0.50\\  \hline

\textbf{SpN} & 	0.50  &	0.36  &	0.40  &	0.12  &	0.34  &	0.29  &	0.25  &	0.36  &	0.24  &	0.38  &	0.37  &	0.45  &	0.30  &	0.41  &	0.38  &	0.40  &	0.38  &	0.47  &	\cellcolor{green}1.00  &	0.29\\  \hline

\textbf{ZB} & 	\cellcolor{yellow}0.61  &	0.25  &	\cellcolor{yellow}0.52  &	0.33  &	\cellcolor{yellow}0.71  &	\cellcolor{yellow}0.59  &	0.45  &	0.44  &	\cellcolor{yellow}0.53  &	\cellcolor{yellow}0.71  &	0.41  &	\cellcolor{yellow}0.57  &	\cellcolor{yellow}0.55  &	\cellcolor{yellow}0.61  &	0.46  &	0.29  &	0.42  &	\cellcolor{yellow}0.53  &	0.31  &	\cellcolor{green}1.00\\

\end{tabular}
}
\caption{Structural Similarity Index Matrix for CIFAR10 corruptions. Legend in Table~\ref{table:abbr-cifar10}. Yellow for 0.5 < SSIM $\leq$ 0.75 and Green for SSIM > 0.75}
\label{table:ssim-cifar10}
\end{table}

% Legend Table
\begin{table}[h!]
\centering
\scalebox{0.7}{
\begin{tabular}{ll}
\toprule
Full Name                & Abbreviation \\
\midrule
In-distribution           & ID           \\
Gaussian Noise           & GN           \\
Brightness               & BR           \\
Contrast                 & CT           \\
Defocus Blur             & DB           \\
Elastic                  & EL           \\
Fog                      & FG           \\
Frost                    & FR           \\
Frosted Glass Blur       & FGB          \\
Gaussian Blur            & GB           \\
Impulse Noise            & IN           \\
JPEG Compression         & JC           \\
Motion Blur              & MB           \\
Pixelate                 & PX           \\
Saturate                 & ST           \\
Shot Noise               & SN           \\
Snow                     & SW           \\
Spatter                  & SP           \\
Speckle Noise            & SpN           \\
Zoom Blur                & ZB           \\
\bottomrule
\end{tabular}
}
\caption{Legend for Abbreviations Used in CIFAR10 corruptions in Table~\ref{table:ssim-cifar10}}
\label{table:abbr-cifar10}
\end{table}

\subsection{Additional insights into debugging}
\label{apdx:debug}
The HDC classifiers are generally outperformed by conventional NNs due to their inherent architecture limitations and binarized implementations. 
Another important factor impacting HDC performance is the presence of \textit{similar corruptions}. Table~\ref{table:ssim-spcmd} and Table~\ref{table:ssim-cifar10} present the structural similarity index measure (SSIM) matrices for Speech Commands (SpCmd) and CIFAR10 corruptions, respectively. These tables indicate the degree of similarity between each pair of corruptions. To compute the SSIM between two corruption types, we randomly sample 50 images from each type and calculate the mean SSIM between them. Each sample pair consists of identical images, each corrupted by a different type of corruption. A SSIM value of 0 indicates no similarity between images, while a value of 1 signifies complete identity. The cells in Tables~\ref{table:ssim-spcmd} and~\ref{table:ssim-cifar10} are color coded for identification of similar corruptions. In these tables, a yellow cell indicates SSIM between 0.5 and 0.75 (moderate similarity), and a green cell indicates SSIM>0.75 (high similarity). As seen in the SSIM table for SpCmd (Table~\ref{table:ssim-spcmd}), there are many corruptions with high degree of similarity, which limits the performance of even a conventional MLP classifier.

The SSIM tables can help developers narrow down the number of corruptions to detect on-device, which in turn leads to a massive accuracy improvement. For example, when we reduce the number of CIFAR10 corruptions from 19 to 12, we immediately improve the top-1 accuracy from 0.68 to 0.79. The SSIM tables help the developers eliminate similar corruption types from the debugging task. For example, in the SSIM table for CIFAR10 (Table~\ref{table:ssim-cifar10}), we observe that defocus blur (DB) and gaussian blur (GB) are highly similar. Therefore, the developer can choose to remove either of them from the classification task to obtain a higher precision in debugging.

% Define color for different ranges
\definecolor{yellow}{rgb}{1.0,1.0,0.6}
\definecolor{green}{rgb}{0.6,1.0,0.6}

\begin{table}[h!]
\centering
\scalebox{0.8}{
\begin{tabular}{l|r|r|r|r|r|r|r|r|r|r|r|r|r}

\textbf{SSIM} & \textbf{ID} & \textbf{GN} & \textbf{AA} & \textbf{BPF} & \textbf{BSF} & \textbf{HPF} & \textbf{HSF} & \textbf{LPF} & \textbf{LSF} & \textbf{PF} & \textbf{TD} & \textbf{TM} & \textbf{TS} \\
\hline
\textbf{ID} & 	\cellcolor{green}1.00  &	0.31  &	\cellcolor{green}0.80  &	0.45  &	0.40  &	0.44  &	\cellcolor{green}0.93  &	\cellcolor{green}0.82  &	\cellcolor{green}0.93  &	\cellcolor{green}0.89  &	\cellcolor{green}0.76  &	\cellcolor{green}0.95  &	\cellcolor{green}0.77\\ \hline

\textbf{GN} & 	0.45  &	\cellcolor{green}1.00  &	0.41  &	0.31  &	0.21  &	0.32  &	0.47  &	0.43  &	0.46  &	0.43  &	0.44  &	\cellcolor{yellow}0.69  &	\cellcolor{yellow}0.59\\ \hline

\textbf{AA} & 	\cellcolor{green}0.77  &	0.24  &	\cellcolor{green}1.00  &	0.48  &	0.38  &	0.43  &	\cellcolor{green}0.80  &	\cellcolor{green}0.84  &	\cellcolor{green}0.78  &	\cellcolor{yellow}0.72  &	\cellcolor{yellow}0.60  &	\cellcolor{green}0.85  &	\cellcolor{yellow}0.67\\ \hline

\textbf{BPF} & 	0.45  &	0.12  &	0.46  &	\cellcolor{green}1.00  &	0.27  &	\cellcolor{yellow}0.55  &	0.46  &	0.46  &	0.42  &	0.41  &	0.33  &	\cellcolor{yellow}0.61  &	0.41\\ \hline

\textbf{BSF} & 	0.40  &	0.09  &	0.35  &	0.25  &	\cellcolor{green}1.00  &	0.24  &	0.34  &	0.33  &	0.35  &	0.38  &	0.29  &	\cellcolor{yellow}0.59  &	0.39\\ \hline

\textbf{HPF} & 	0.40  &	0.15  &	0.44  &	\cellcolor{yellow}0.55  &	0.27  &	\cellcolor{green}1.00  &	0.46  &	0.45  &	0.47  &	0.49  &	0.29  &	\cellcolor{yellow}0.65  &	0.47\\ \hline

\textbf{HSF} & 	\cellcolor{green}0.94  &	0.31  &	\cellcolor{green}0.77  &	0.48  &	0.39  &	0.45  &	\cellcolor{green}1.00  &	\cellcolor{green}0.82  &	\cellcolor{green}0.90  &	\cellcolor{green}0.86  &	\cellcolor{yellow}0.71  &	\cellcolor{green}0.93  &	\cellcolor{green}0.79\\ \hline

\textbf{LPF} & 	\cellcolor{green}0.85  &	0.26  &	\cellcolor{green}0.82  &	0.44  &	0.36  &	0.38  &	\cellcolor{green}0.83  &	\cellcolor{green}1.00  &	\cellcolor{green}0.81  &	\cellcolor{yellow}0.73  &	\cellcolor{yellow}0.61  &	\cellcolor{green}0.86  &	\cellcolor{yellow}0.65\\ \hline

\textbf{LSF} & 	\cellcolor{green}0.93  &	0.26  &	\cellcolor{green}0.77  &	0.43  &	0.35  &	0.47  &	\cellcolor{green}0.89  &	\cellcolor{green}0.84  &	\cellcolor{green}1.00  &	\cellcolor{green}0.83  &	\cellcolor{yellow}0.69  &	\cellcolor{green}0.93  &	\cellcolor{yellow}0.74\\ \hline

\textbf{PF} & 	\cellcolor{green}0.87  &	0.25  &	\cellcolor{green}0.78  &	0.46  &	0.37  &	0.46  &	\cellcolor{green}0.85  &	\cellcolor{yellow}0.73  &	\cellcolor{green}0.81  &	\cellcolor{green}1.00  &	\cellcolor{yellow}0.69  &	\cellcolor{green}0.91  &	\cellcolor{yellow}0.72\\ \hline

\textbf{TD} & 	\cellcolor{green}0.80  &	0.31  &	\cellcolor{yellow}0.64  &	0.38  &	0.31  &	0.38  &	\cellcolor{green}0.78  &	\cellcolor{yellow}0.71  &	\cellcolor{yellow}0.73  &	\cellcolor{yellow}0.71  &	\cellcolor{green}1.00  &	\cellcolor{green}0.85  &	\cellcolor{yellow}0.66\\ \hline

\textbf{TM} & 	\cellcolor{green}0.90  &	0.29  &	\cellcolor{yellow}0.73  &	0.44  &	0.37  &	0.40  &	\cellcolor{green}0.85  &	\cellcolor{green}0.76  &	\cellcolor{green}0.84  &	\cellcolor{green}0.80  &	\cellcolor{yellow}0.72  &	\cellcolor{green}1.00  &	\cellcolor{yellow}0.74\\ \hline

\textbf{TS} & 	\cellcolor{yellow}0.69  &	0.24  &	\cellcolor{yellow}0.56  &	0.37  &	0.32  &	0.32  &	\cellcolor{yellow}0.69  &	\cellcolor{yellow}0.60  &	\cellcolor{yellow}0.64  &	\cellcolor{yellow}0.67  &	\cellcolor{yellow}0.56  &	\cellcolor{green}0.85  &	\cellcolor{green}1.00\\

\end{tabular}
}
\caption{Structural Similarity Index Matrix for SpeechCmd corruptions. Legend in Table~\ref{table:abbr-spcmd}. Yellow for 0.5 < SSIM $\leq$ 0.75 and Green for SSIM > 0.75}
\label{table:ssim-spcmd}
\end{table}

\begin{table}[h!]
\centering
\scalebox{0.7}{
\begin{tabular}{c|c}
\hline
\textbf{Abbreviation} & \textbf{Corruption type} \\
\hline
ID & In-distribution  \\
GN & Gaussian Noise \\
AA & Air Absorption \\
BPF & Band Pass Filter \\
BSF & Band Stop Filter \\
HPF & High Pass Filter \\
HSF & High Shelf Filter \\
LPF & Low Pass Filter \\
LSF & Low Shelf Filter \\
PF & Peaking Filter \\
TD & Tanh Distortion \\
TM & Time Mask \\
TS & Time Stretch \\
\hline
\end{tabular}
}
\caption{Legend for Abbreviations Used in SpeechCmd corruptions in Table~\ref{table:ssim-spcmd}}
\label{table:abbr-spcmd}
\end{table}

\subsubsection{Effect of removing corruptions from debugging}
Demonstrating the usability of a debugging tool is a complex task, as each developer may use it differently. We attempt to explore a plausible debugging scenario, where corruptions are removed from the HDC classification task to improve accuracy.

While removing corruptions from the classification may potentially reduce the coverage of the debugging task, we find that this might not always be the case. For example, consider a scenario where Gaussian blur (GB) is removed from the classification task, but defocus blur (DB) remains. Post-deployment, if the input is corrupted by GB, the HDC classifier might still identify it, though misclassifying it as DB. In response, the developer may choose to retrain the model to handle DB corruption. Interestingly, this retraining phase may indirectly improve the model's ability to handle GB corruption as well, due to the similarity between DB and GB. We empirically verify this claim by showing that after retraining on DB, the model exhibits improved performance on GB-corrupted inputs, even though GB was not explicitly included during retraining. This is also supported by the findings of~\citet{pieter2018-patternet}, where they developed an \textit{explanation} for neural networks. According to~\citet{pieter2018-patternet}, input data consists of two components: the \textit{signal}, which contains task-relevant information, and the \textit{distractor}, which obfuscates the signal. During training, neural networks learn to filter out the distractor to recover the signal.
Applied to our scenario, if the network learns to remove DB to recover the signal, it should similarly be able to remove GB. Thus, developers can safely exclude similar corruptions from the classification task, improving HDC accuracy with minimal impact on the debugging process.

\end{document}